\documentclass[10pt,twocolumn,letterpaper]{article}

\usepackage{isba}
\usepackage{times}
\usepackage{epsfig}
\usepackage{graphicx}
\usepackage{amsmath}
\usepackage{amssymb}
\usepackage{url}
\usepackage[linesnumbered,ruled,vlined]{algorithm2e}

\isbafinalcopy 

\ifisbafinal\fi
\begin{document}

\title{ FDFNet : A Secure Cancelable Deep Finger Dorsal Template Generation Network Secured via. Bio-Hashing}
\author{Avantika Singh\\
IIT Mandi\\
Mandi, India\\
{\tt\small d16027@students.iitmandi.ac.in}
\and
Ashish Arora\\
IIT Dharwad\\
Dharwad, India\\
{\tt\small 160010008@iitdh.ac.in}
\and
Shreya Hasmukh Patel\\
IIT Jodhpur\\
Jodhpur, India\\
{\tt\small hasmukh.1@iitj.ac.in}
\and
Gaurav Jaswal\\
IIT Mandi\\
Mandi, India\\
{\tt\small gaurav\_jaswal@projects.iitmandi.ac.in}
\and
Aditya Nigam\\
IIT Mandi\\
Mandi, India\\
{\tt\small aditya@iitmandi.ac.in}
}

\maketitle
\thispagestyle{empty}


\begin{abstract}
   Present world has already been consistently exploring the fine edges of online and digital world by imposing multiple challenging problems/scenarios. Similar to physical world, personal identity management is very crucial in-order to provide any secure online system. Last decade has seen a lot of work in this area using biometrics such as face, fingerprint, iris etc. Still there exist several vulnerabilities and one should have to address the problem of compromised biometrics much more seriously, since they cannot be modified easily once compromised. In this work, we have proposed a secure cancelable finger dorsal template generation network (learning domain specific features) secured via. Bio-Hashing. Proposed system effectively protects the original finger dorsal images  by withdrawing compromised template and reassigning the new one.  A novel Finger-Dorsal Feature Extraction Net (FDFNet) has been proposed for extracting the discriminative features. This network is exclusively trained on trait specific features without using any kind of pre-trained architecture. Later Bio-Hashing, a technique based on assigning a tokenized random number to each user, has been used to hash the features extracted from FDFNet. To test the performance of the proposed architecture, we have tested it over two benchmark public finger knuckle datasets: PolyU FKP and PolyU Contactless FKI. The experimental results shows the effectiveness of the proposed system in terms of security and accuracy.
\end{abstract}

\section{Introduction}
\begin{figure}[!htp]
	\begin{center}
		\includegraphics[width=1\linewidth,height=0.56\linewidth]{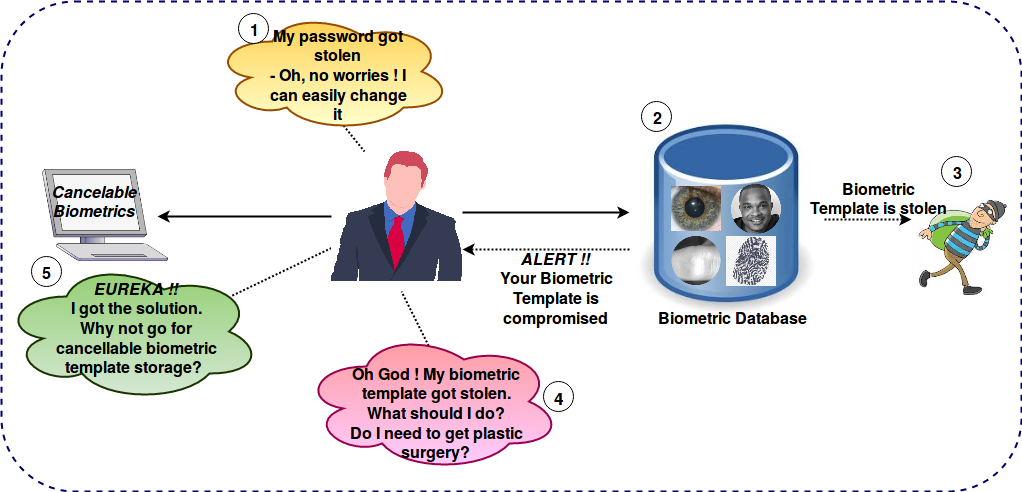}
		
	\end{center}
	\caption{Need of cancelable biometrics}
	\label{fig:start}
\end{figure}

Due to the tremendous growth in automated secured systems; the security and privacy of personal information has become one of the most critical and challenging task of the present day world. Even the most common personnel security mechanisms like password or PINs, card and biometrics have appeared to be insufficient in addressing the challenges of identity frauds. Passwords or PINs are easily stolen and forgotten, cards are easily duplicated and lost whereas biometric traits suffer from the privacy assault and non-revocable issues. It simply means, if a biometric is compromised, it is rendered worthless, just like a password or PINs. In such circumstances, replacement of a biometric trait with new template is not reasonable because biometrics is permanently associated with an individual and cannot be replaced even if compromised. Thus, to protect biometric data from external attacks or privacy invasion, the concept of cancelable biometrics has been introduced as depicted in Fig.\ref{fig:start}.
In particular, cancelable biometrics is a feature transformation where instead of storing the original biometric, it is transformed using a non-reversible function. In this way, it is computationally difficult to reconstruct the original biometric from a transformed one. Several biometric template protection schemes have been well proposed in the literature. However in terms of comparison, biometric solutions are more reliable and successful for authenticating an individual than passwords or card based mechanisms because they carry unique biological information of an individual. Hence, due to the usability merit, biometrics are increasingly replacing conventional PINs, cards in many modern security applications.
One of the most popular biometric traits used in security systems are fingerprint, iris and face. Unfortunately, they have several trait specific drawbacks that pose limitation to their usages in low level security applications. For instance, occlusion and reflection affects the iris, poor quality and social acceptance related issues for fingerprint, as well as pose, expression, and aging for face recognition. Nevertheless, the good performance of finger knuckle print represents a recent trend in this field.

 In literature, most of the studies \cite{deep_knuckle,rknuckle} are based on major finger-knuckle-print (FKP) as a biometric-trait. But recently, several papers have reported systems using upper and lower minor finger knuckle \cite{rdorsal} and nail \cite{rnail}  that have certainly shown great potential to augment FKP for better performance. In this work, we have examined major finger knuckle, upper minor knuckle and nail which is together coined as  \emph{finger dorsal image} in the literature \cite{rdorsal} just as a novel case study, majorly because its  cancellability has not been studied at all. However, we have experimentally observed, that the proposed technique is equally applicable to other biometric traits like iris, face or fingerprints. 
The convex shape lines and creases on finger dorsal surface are very distinctive and permanent to everyone and are easily collectible using low-resolution imaging cameras. These features are invariant to emotions, aging effects and are naturally preserved to injuries and excessive usages (unlike fingerprints) because they exist on the outer side of finger. With such tremendous benefits, it motivates us to work over the problem of finger knuckle template protection.

\textbf{Related Work:} The security and privacy issues of biometric templates have been emerged as an important problem, especially when the databases are centrally accessed or networked. The various approaches on biometric template protection have been devised since last 18 years. Out of which, non-invertible transformation methods are more robust and maintain recognition performance too. In this context, the first ever work on biometric template security was given by \cite{1}, in which they discussed some important aspects of biometrics security threats and privacy invasion. In another study \cite{3}, the authors defined non-invertible transformation for constructing fingerprint templates. In \cite{2}, the fingerprint minutiae based rotation and translation invariant features were used to generate alignment free cancelable templates. Likewise in \cite{4}, a registration free method for constructing cancelable fingerprint templates by using non-irreversible transformation known as curtailed circular convolution was introduced. In \cite{5}, authors presented an irreversible representation of fused features of face and iris biometrics to improve the privacy protection compared to a single feature. In another work \cite{6}, a cancelable iris recognition system based on feature learning and 2 stage Bio-Hashing scheme was developed. With no pre-alignment \cite{7}, authors proposed to exploit zoned minutia feature into  non-invertible fingerprint representation by using partial DFT(Discrete Fourier transform). In \cite{10}, authors presented a cancelable multi-biometric systems in which fused features of fingerprint and finger vein are transformed into common template by using enhanced partial discrete Fourier transform. To the best of our knowledge \cite{9} is the only paper in which CNN is used to learn robust binary representation of deep face features which are further hashed using cryptographic method known as SHA-3 $512$. 

\subsection{Major Concerns and Contribution} 
The major open issues in knuckle biometrics are the lack of robustness against outdoor illumination, low image quality, inconsistent ROI segmentation and poor matching between weaker texture regions. 
To the best of our knowledge, this is the first time that efforts have been  made to introduce finger dorsal  modality as cancelable biometric template, where the CNN (proposed FDFNet) based domain specific biometric features of a finger dorsal modality are transformed using a well known Bio-hashing technique \cite{hash}. Apart from major and minor finger knuckles, we have worked over nail which has not been used till now. The matching performance of our method revealed that the obtained bio-hashed finger dorsal templates have properties like non-invertibility, diversity, and revocability to be consider as an ideal cancelable biometric template. This work can be regarded as a preliminary work for investigating the potential of deep-learning in generating cancelable templates.
\begin{figure}[!htp]
	\begin{center}
		\includegraphics[width=0.99\linewidth,height=0.40\linewidth]{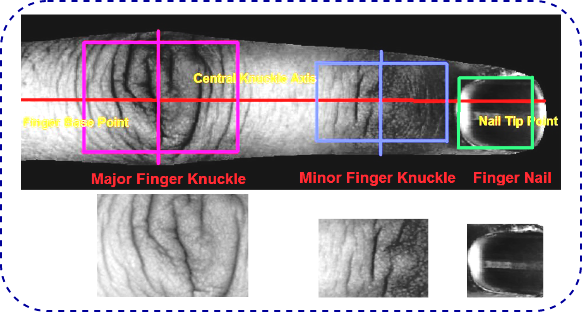}
		
	\end{center}
	\caption{Finger knuckle image annotation (Sample image from PolyU FKI dataset)}
	\label{fig:fki}
\end{figure}

\begin{figure*}[!htp]
	\begin{center}
		\includegraphics[width=17cm,height=17 cm,keepaspectratio,]{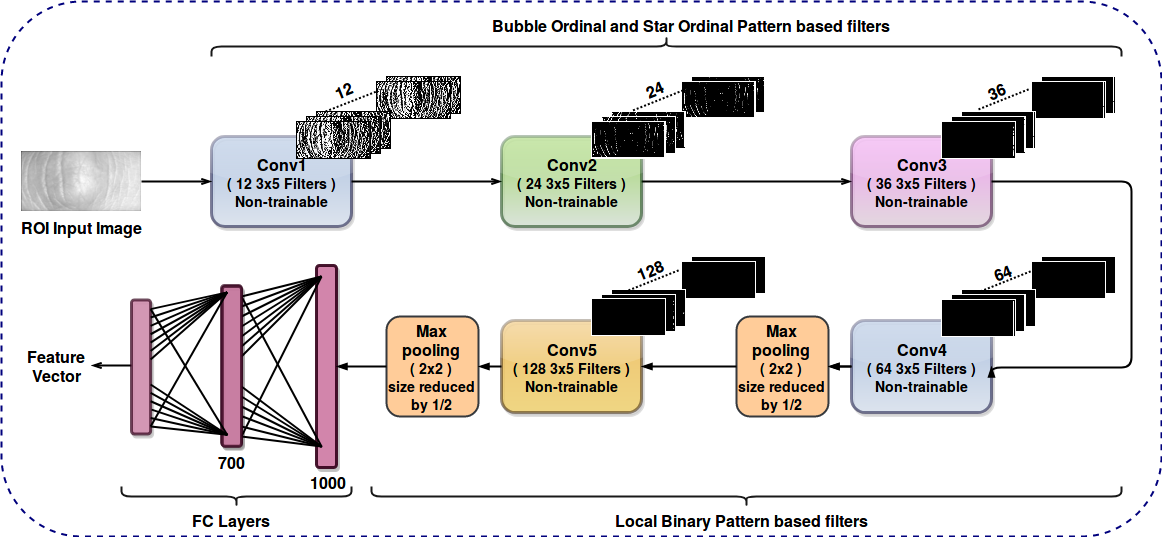}\\
		
	\end{center}
	\caption{ Finger dorsal feature extraction network (FDFNet) to learn domain specific features}
	\label{fig:net}
\end{figure*}

 \textbf{Model Justification :} In this problem, we cannot train a deeper CNN because we don't have a large training data of finger dorsal samples. Therefore, we utilized the domain knowledge to learn finger dorsal domain specific features, so that we can get most discriminative knuckle features. In proposed FDFNet, most of the filters in convolution layers are fixed (non-trainable) so as to learn domain knowledge as much as possible. This will lead to lesser trainable parameters only for fully connected layers. On the other hand, if we select pre-train VGG and ResNet models to extract features, then their fine tuning was very difficult and they got over-fitted very soon. Hence, we have proposed a light CNN model known as FDFNet. This model learns trait specific features  using CNN filters based on local adaptive procedures such as LBP (Local Binary Patterns), BOP (Bubble ordinal pattern) \cite{deep_knuckle} etc., to improve the textural strength and handle illumination variations.
 
 The remainder of the article is organized into following five main sections: Section 2 describes the proposed cancelable biometric system which include feature extraction n/w, data augmentation, template generation and transformation. In section 3. the experimental results are presented. Next, we discussed on the various aspects of security analysis. Finally, conclusion is drawn in last section.
 
 \begin{figure*}[!htp]
 	\begin{center}
 		\includegraphics[width=0.99\linewidth, height=0.55\linewidth]{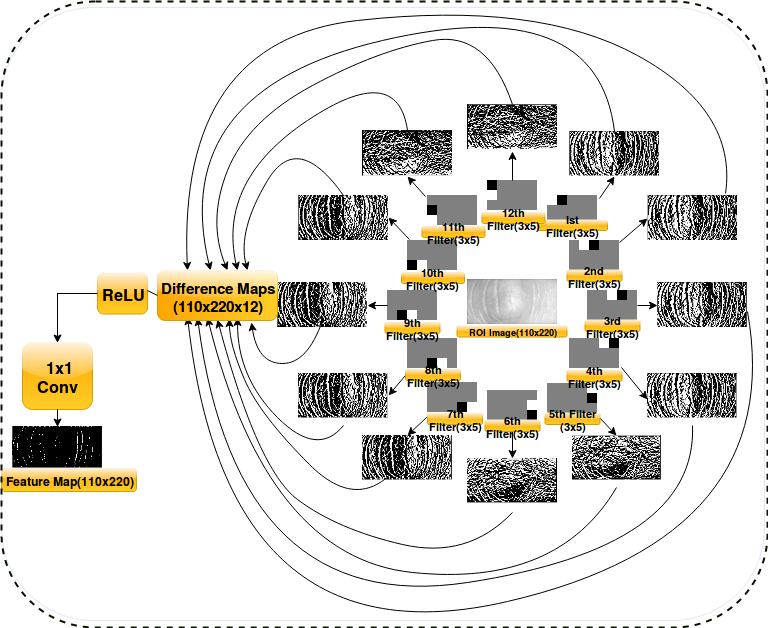}\\
 		
 	\end{center}
 	\caption{ $1^{st}$ Layer FDFNet based feature map representation}
 	\label{fig:feature}
 \end{figure*}
 \section{Proposed Methodology} With small size databases in hand it is very cumbersome to learn the entire network from scratch without using any kind of pre-trained architecture or model. At first, we have extracted ROI's of major knuckle, minor knuckle and nail traits over PolyU FKI dataset by training a state-of-the-art region based convolutional neural network (CNN) that use different bounding boxes as ground truth to classify and localize the ROI's \cite{11}.
  Performance of this network, for segmenting images in terms of accuracy, at an overlap IOU threshold of $0.5$ is as follows:
  
  \textemdash PolyU FKP-Major \cite{15}: $98.92\%$ 
  
  \textemdash PolyU FKI-Major, Minor, Nail \cite{16}: $99.63\%$, $89.94\%$, $98.48\%$ respectively. Since the erroneous segmentation can severely affects the feature extraction and identification performance, we have manually segmented erroneously segmented images in our case. This helps us to conclusively investigate our claim of security and cancellability without addressing the other major issues such as segmentation or image quality. An example of an annotated finger dorsal image is shown in Fig.\ref{fig:fki}. For extracting the distinctive features from the obtained ROI's dataset, we have trained a CNN architecture from scratch. Since we have limited amount of dataset with us, training a deep network that requires lot of parameters to be learned is quite tedious. In order to escape this situation, we have proposed STAR ordinal pattern(SOP) and Bubble ordinal pattern (BOP) \cite{deep_knuckle} based network in conjunction with local binary convolution, an effective surrogate to convolutional layers in standard convolutional neural networks(CNN). This network is inspired from \cite{lbp_cnn}. Our proposed network architecture is shown in Fig.\ref{fig:net}.

  The main reason for choosing this type of network is that all the three patterns SOP, BOP and LBP are quite robust for illumination variance and more-over knuckle data is having quite good textual patterns that are well captured by  them. Secondly, the LBC(Local binary Convolutional) layer comprises of a set of fixed sparse pre-defined binary convolutional filters that are not trainable similarly SOP and BOP filters are not updated during the training process. This significantly reduces the number of the learn-able parameters in our network and thus we are able to train our network with less data from scratch and that too without over-fitting. We have trained this network in such a manner that it is able to learn domain specific finger dorsal textural features.

 \begin{figure}[!htp]
 	\begin{center}
 		\includegraphics[width=0.99\linewidth, height=0.48\linewidth]{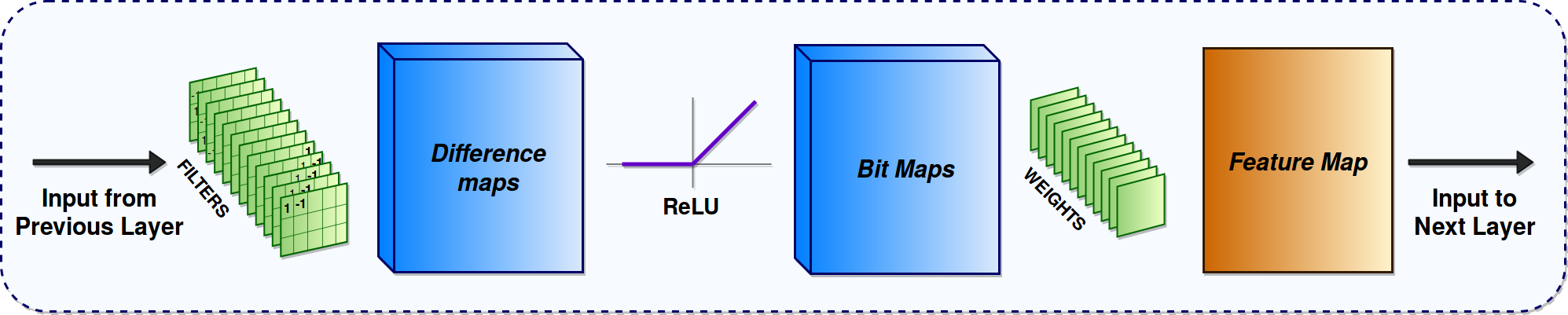}\\
 		
 	\end{center}
 	\caption{ Bit Map and difference map generation in FDFNet architecture}
 	\label{fig:bitmap}
 \end{figure}
 
 \textbf{Finger dorsal feature extraction network (FDFNet) architecture:} 
 \begin{enumerate}
 	\item The first layer of our convolutional FDFNet is shown in Fig.\ref{fig:feature}. It contains $12$ short and wide filters having dimensions $3\times5$. The main motive of using these tall and thin filters is the aspect ratio of our image which is $0.5$. The weights of these filters are non-trainable, rather they are initialized in a specific pattern. Each convolution filter is a difference filter having $1$ and $-1$ initialized on the adjacent boundary boxes and zeros at all other places. In total, $12$ filters are obtained with this kind of distribution. These are also known as Bubble Ordinal Pattern filters. After convolving with these $12$ filters, a $12$ channel output is obtained. Each output channel is known as a difference map, which on further application of ReLU as a non-linear activation function, gives a bit-map as shown in Fig.\ref{fig:bitmap}. Subsequently, a $1 \times 1$ convolution is applied to obtain the weighted sum of all the bit maps, generating a single channel output which is input to the second convolutional layer. The weights of these $1 \times 1$ filters are trainable.
 	
 	\item The second convolutional layer consists of $24$ difference filters, in which first $12$ filters are initialized with $1$ and $-1$ with a gap of $2$ and the remaining $12$ filters are initialized with a gap of $3$.

 	\item Third convolutional layer consists of $36$ filters and each $12$ filters are initialized with a gap of $4$, $5$ and $6$. The single channel output of the third convolutional layer is passed onto the fourth convolutional layer which is a LBC layer consisting of $64$ difference filters. The weights of these filters are also non-trainable. 
 	\item In a LBC layer, filters weights are randomly initialized with $1$,$-1$ and $0$. Sparsity and Bernoulli distribution are $2$ factors which helps in governing the number of zeros and ones in a filter. Sparsity indicates the fraction of the zeros in the filter and Bernoulli distribution determines the distribution of $1’s$ and $-1’s$. The $64$ channel output is convolved with $64$ ($1\times1$ filters), to obtain a single channel output. Subsequently, max pooling of $2\times2$ is applied to reduce the dimension of the feature map by half.
 	
 	\item The fifth convolutional layer is also a LBC layer comprising of $128$ difference filters and $2\times2$ pooling. After convolving with these $128$ ($3\times5$) filters and taking the weighted sum, max pooling is applied to obtain the output feature map. This output feature map is flattened into a single dimension feature vector which is given as the input to the remaining $3$ fully connected layers. The dimension of the last FC layer depends upon the number of the subjects present in the dataset.  
 	
 \end{enumerate}
 
  We have designed FDFNet by logically justifying and experimentally verifying each and every network component. To achieve best filter response three sets of experimentation with varying filter sizes have been  performed over PolyU FKP [18] dataset: (i) $3\times3$ (square filter), (ii) combination of $3\times3$ and $3\times5$ filters and (iii) $3\times5$ (rectangular filter) which results in (train, test) accuracy as ($85\%$, $78\%$), ($95\%$, $79\%$), ($95\%$, $87\%$) respectively.
  
 \subsection {Data Augmentation} Data augmentation is prerequisite in our case since we have limited amount of training samples at our disposal and convolutional neural networks require large amount of training data for discriminative information learning. We perform data augmentation on each input image using Augmentor library \cite{augmentor} of python with operations like zooming, random distortion, illumination variation and rotation. For each input image a total of $45$ augmented images are generated.
 \subsection{ Cancelable finger dorsal template generation } 
\begin{figure}[!htp]
 	\begin{center}
 		\includegraphics[width=1\linewidth, height=0.80\linewidth]{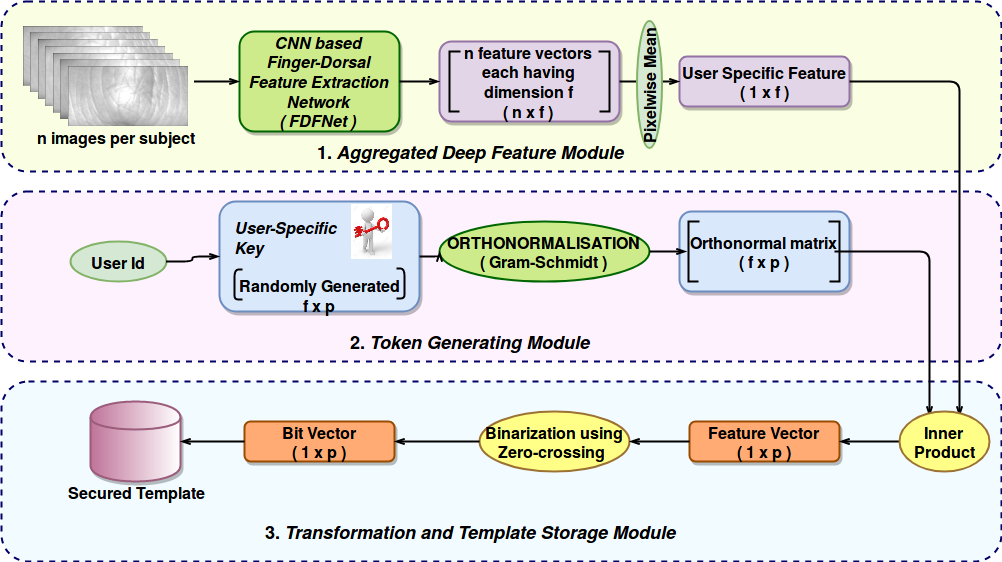}\\
 		
 	\end{center}
 	\caption{ Multishot enrollment architecture for Cancelable finger dorsal biometric system}
 	\label{fig:multishot}
 \end{figure}

Multishot enrollment architecture for cancelable finger dorsal biometric system is depicted in Fig.\ref{fig:multishot}. It is broadly divided into three subparts. The first part (Aggregated deep feature module) mainly deals with extracting finger dorsal specific features. FDFNet as described in previous section is used to extract these domain specific features. The second subpart user token generation module is used to assign a unique identifier to each subject while the third part deals with generating cancelable template and its storage. Since, we are using a deep CNN for extracting the features we go for multishot enrollment in order to train the network and to learn robust feature representation. We take the mean representation of all the features corresponding to a single subject.
 \subsection{Transformation}
 \label{sec:trans} 
 
 One of the important property of the transformation process is to transform the feature vector into a new space such that the intra class similarity and inter class dissimilarity is preserved. For this we have used state-of-the-art Bio-Hashing technique \cite{hash}. This process is a four step process as follows:
 \\
 
 \textbf{Step 1:} A set of pseudo-random vectors are generated that acts as a user-specific key and stored in the database (used during verification process) as:
 \begin{equation}
 r_i \in  R^M | i=1,---,m
 \end{equation}
 
 \textbf{Step 2:} Orthogonalization of pseudo-random vectors in inner product space is done using Gram-Schmidt process as:
 \begin{equation}
 r_{\perp i} \in  R^M | i=1,---,m
 \end{equation}
 
 \textbf{Step 3:} Next step is to project the feature vectors $f_i$---$f_n$ (say for n subjects) computed using FDFNet on orthonormal basis computed in Step 2 by calculating the inner product as:
 
 \begin{align} \label{eq:3}
 \left[f_1......f_n\right]\left[ {\begin{array}{ccc}
 	r_{11} & ..........&r_{1m} \\
 	.......& .........&...... \\
 	...... & .........&....... \\
 	r_{1n} & .......&r_{nm} \\
 	\end{array} } \right]=\left[B_1....B_m\right]
 \end{align}
 
 \textbf{Step 4:} The resultant feature vectors obtained in Step 3 are real value feature vectors which are further thresholded using Zero-crossing technique as :
 \begin{equation} \label{eq:4}
 Q_i=\begin{cases}
 1, & \text{if $B_i>0$}.\\
 0, & \text{otherwise}.
 \end{cases}
 \end{equation}

 \subsection{Verification Architecture for Cancelable Finger Dorsal Biometric System}
\begin{figure}[!h]
 	\begin{center}
 		\includegraphics[width=1\linewidth,height=0.6\linewidth=0.35]{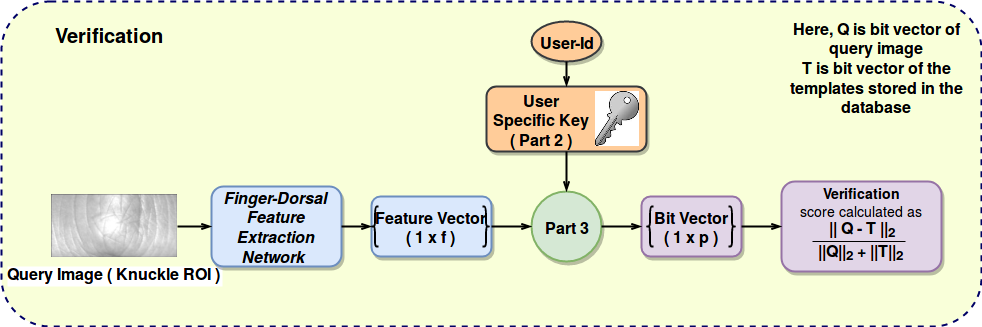}\\
 		
 	\end{center}
 	\caption{ Verification architecture for cancelable finger dorsal biometric system } 
 	\label{fig:veri}
 \end{figure}

 Once the cancelable template is stored in the database, the next step is to verify the claim identity of the user at the time of verification. The steps involved in this process are as follows:
 
 \textbf{Step 1:} First the query image  belonging to subject $i$ say $image_i$ is passed to same FDFNet architecture to extract deep features say $f_i$ as shown in Fig.\ref{fig:veri}.

 \textbf{ Step 2:} Once the feature vector $f_i$ is obtained,  as discussed in transformation section the  vector $B_i$ corresponding to query image $image_i$ is generated using Equation \ref{eq:3} and further binarized using Zero-crossing to generate vector $Q_i$ as discussed  in Equation \ref{eq:4}.

 \textbf{ Step 3:} The last step is to calculate the similarity measure between the resultant bit-vector and the stored bit-vectors in our database  to verify the claim identity. The metric used for similarity measure  is as follows:
 \begin{equation}
 \frac{\lVert {Q_i-T} \rVert_{2} }{{\lVert {Q_i} \rVert}_{2}+ {\lVert {T} \rVert}_{2} }
 \end{equation}
 
 Here $Q_i$ is the bit vector corresponding to query image in our case and $T$ is the bit vector corresponding to different  templates stored in our database. We have used this metric because most of the earlier proposed studies on cancelable biometrics had already used this type of  L2 similarity metrics \cite{metric}. Here in Fig.\ref{fig:veri} the Part 2 of the diagram is considered with binding the user-specific key with the biometric features extracted using FDFNet while Part 3 is considered with calculating the inner product between feature vectors and the orthonormal feature vectors extracted from user-specific key. Verification algorithm that verifies the claim identity of the query image is given in algorithm \ref{algo:b} with detailed steps.
 \begin{algorithm}
 	\KwIn{input image \textbf{q}, claim identity \textbf{i}}
 	\KwOut{score s, probability by which q $\notin$ i}
 	For the query image \textbf{q} first extract the ROI corresponding to \textbf{q} say \textbf{$q_{ROI}$}.\\
 	Pass \textbf{$q_{ROI}$} to FDFNet to extract features corresponding to it say \textbf{$q_f$} and similarly pass the template of the claimed identity \textbf{i}  to  FDFNet to extract features say \textbf{$i_f$} .\\
 	Generate random vector corresponding to the claim identity say \textbf{$r_i$}.\\
 	Apply Gram Schmidt to \textbf{$r_i$} to generate orthonormal basis say \textbf{$r_{\perp i}$}. \\
 	Compute the inner product between \textbf{$q_f$} and 
 	\textbf{$r_{\perp i}$} , similarly compute the inner product between \textbf{$i_f$} and 	\textbf{$r_{\perp i}$}. Let it be \textbf{Q} and \textbf{$T_i$} respectively.\\
 	Calculate the score between \textbf{Q} and \textbf{$T_i$} as:
 	\begin{equation}
 	\frac{\lVert {\mathbf{Q}-T_i} \rVert_{2} }{{\lVert \mathbf{Q} \rVert}_{2}+ {\lVert {T_i} \rVert}_{2} }
 	\end{equation}
 	Score closer to 1 represents dissimilarity from the claim identity while score closer to 0 verify the claim identity.\\

 	\caption{Verification Algorithm}
 	\label{algo:b}
 \end{algorithm}
  \subsection{Network Implementation Details} The proposed network has been implemented using python and tensorflow\cite{tensorflow} library. All the indigenous experiments have been done on the same platform as Intel(R) Xenon(R) CPU E5-2630 V4 at 2.20 GHz with 32 GB RAM and NVIDIA Tesla K40 C GPU card with 12 GB on card RAM.
  \section{Experimental Analysis and Discussions}
  
  \begin{figure}[!htp]
  	\small
  	\centering
  	\includegraphics[width=1\linewidth,height=0.8\linewidth]{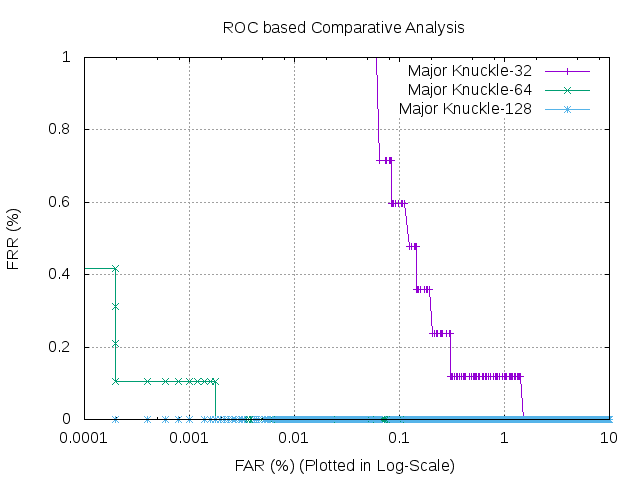} %
  	\caption{ROC based performance analysis for major knuckle cancelable biometrics}
  	\label{fig:a1}
  \end{figure}
  In distinguishing experiments, the performance of proposed approach has been evaluated in terms of EER (Equal Error Rate), CRR (Correct Recognition Rate) and DI (Decidability Index). We also show variants of EER like FAR (False Acceptance Rate) and FRR (False Rejection Rate) for rigorous testing. The performance metrics are computed based on the Euclidean distance during matching with 3 different bit maps i.e., 32 bit, 64 bit and 128 bit.\\
  \textbf{Database Specifications :} The two publicly available finger knuckle databases i.e., PolyU FKP \cite{15}, and PolyU Contactless FKI \cite{16} have been used for the analysis and validation of proposed approach. However, the images also suffer from finger artifacts, low contrast, illumination variation, reflection, and nonrigid deformations. The PolyU FKI consists of 5 finger images per subject for all the 503 classes. While, the 6 left and 6 right finger samples from a subject are considered belonging to separate individuals, which resulted in 660 subjects. It is to be noted that, the PolyU FKI dataset \cite{16} has a complete finger dorsal images over which major knuckle, minor knuckle and nail ROI's are extracted. While, PolyU FKP dataset \cite{15} only provides major finger knuckle images.
  
  \textbf{Testing Protocol :} Since both of the above discussed databases have different images and class numbers. During testing on PolyU FKP, the first six samples per subject are considered as gallery and the remaining six as probe. This involves $3,960$ genuine matchings and $26,09,640$ impostor matching scores. On the other hand in case of PolyU FKI database, first 3 samples per subject are selected for training and rest two for testing. Therefore, we obtain $1,509$ genuine matching scores and $7,57,518$ impostor matching scores.
  
\begin{figure}[!htp]
  	\small
  	\centering
  	\includegraphics[width=1\linewidth,height=0.8\linewidth]{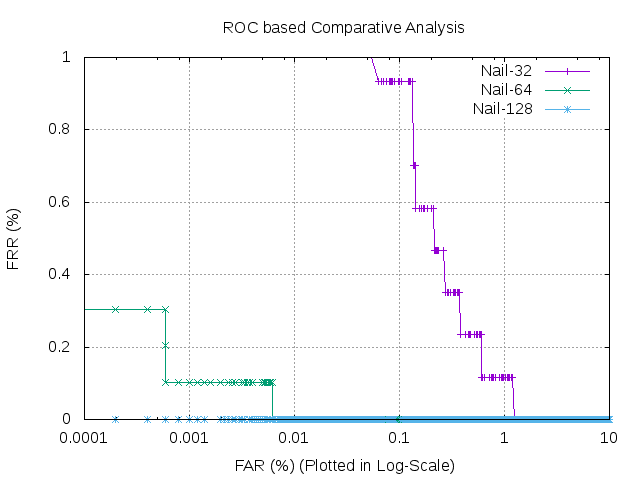} %
  	\caption{ROC based Performance Analysis for Nail cancelable Biometrics}
  	\label{fig:a3}
  \end{figure}

  \textbf{Experiment-1 :} The experimental results are shown in Table \ref{tab:1x}. We report EER, CRR (rank-1) and DI for two datasets with multishot enrollment. Three type of binary bit maps with dimensions d= $32, 64, 128$ are selected for all studied cases. The Receiver Operating Characteristic (ROC) for Major Finger Knuckle and Nail cancelable biometrics are shown in Figure \ref{fig:a1} and Figure \ref{fig:a3} respectively. During testing on $32$ bit map PolyU FKI database, we achieve comparative performance on major finger knuckle and minor finger knuckle while some what lower performance on nail biometric.

  \begin{table}
  	\centering
  	\caption{Summarized results for cancelable finger knuckle biometrics}
  	\label{tab:1x}
  	\begin{tabular}{|p{3cm}|p{1.5cm}|p{0.8cm}|p{0.7cm}|p{0.38cm}|}
  		\hline
  		\textbf{Description}   & \textbf{FAR/FRR}   & \textbf{EER} (\%)  & \textbf{CRR} (\%)  & \textbf{DI}  \\ \hline
  		\multicolumn{5}{|c|}{\textbf{FKI Matching}: 3 Training/ 2 Testing (PolyU FKI Data)} \\ \hline
  		Major Knuckle (32 bit)   &   0.206/0.198   &     0.203      & 99.80       & 2.98 \\ \hline
  		Major Knuckle (64 bit)   &  0.002/0     & 0.0008  & 100        &  4.39   \\ \hline
  		Major Knuckle (128 bit)   & 0/0        & 0     & 100  &6.30   \\ \hline
  		Minor Knuckle (32 bit)   &    0.198/0.198                      &     0.1984    & 100      & 3.04\\ \hline
  		Minor Knuckle (64 bit)   &   0.004/0     & 0.002  & 100         &  4.48  \\ \hline
  		Minor Knuckle (128 bit)   &  0/0        & 0     & 100  & 6.30    \\ \hline
  		Nail(32 bit)   &   0.303/0.298   &     0.300     &99.80       & 3.02  \\ \hline
  		Nail (64 bit)   &  0.006/0.101     & 0.003  &100        &  4.32   \\ \hline
  		Nail (128 bit)   & 0/0 & 0     & 100 & 5.75    \\ \hline
  		\multicolumn{5}{|c|}{\textbf{FKP Matching}: 6 Training/ 6 Testing (PolyU FKP Data )} \\ \hline
  		Major FKP (32 bit)   &  0.071/0.101   &     0.089      &100       & 2.92   \\ \hline
  		Major FKP (64 bit)   & 0/0.02    & 0 & 100         &  4.40   \\ \hline
  		Major FKP(128 bit)   & 0/0   & 0    & 100  & 6.01    \\ \hline
  		\multicolumn{5}{|c|}{\textbf{Finger Components Fusion}: (PolyU FKI Data )} \\ \hline
  		Fused-1 (32 bit)   & 0.002/0.09  &     0.001      &100       & 3.94   \\ \hline
  		Fused-2 (64 bit)   &   0/0.099 &     0    &100       & 5.70   \\ \hline
  		Fused-3  (128 bit)   & 0/0   &     0      &100     & 7.88   \\ \hline
  	\end{tabular}
  \end{table}
  In case of 64 bit map testing, we achieve a similar trend in overall recognition results. An EER of $0.0008 \%$ and CRR of $100 \%$ are achieved on major finger knuckle templates, which are still better than minor finger knuckle (ERR of $0.002$ \%) and nail (ERR of $0.003$ \%). Likewise, the performance difference can be seen in terms of DI between different bit maps. 
  
\begin{figure}[!htp]
  	\small
  	\centering
  	\includegraphics[width=1\linewidth,height=0.8\linewidth]{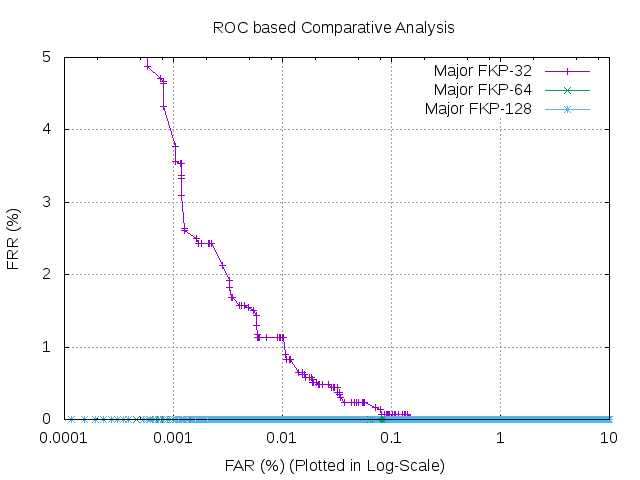} %
  	\caption{ROC based performance analysis for FKP cancelable biometrics}
  	\label{fig:a4}
  \end{figure}

  Further, it is to be noted that the proposed approach has surpassed all the previous cases, when tested on 128 bit maps. One can see higher recognition results with 128 bit map i.e., EER of $0\%$ and CRR of $100 \%$ on each type of finger knuckle template. In addition, the performance of major finger knuckle print on PolyU FKP database is also getting improved as we consider high dimension (32 bit to 128) bit maps. The ROC plot for Major Finger Knuckle Print cancelable biometric is shown in Figure \ref{fig:a4}. From all the tests performed in experiment 1, it can be observed from Table \ref{tab:1x} that: (1) In overall, major knuckle is observed to be a best performing encoded template among nail and finger minor. (2) Nail, which is not studied much in literature is also proved to have better shape and textural features than minor finger knuckle. The performance of proposed approach goes lower to higher w.r.t high dimension bitmaps as the number of the available bits increased.
  
  \textbf{Experiment-2 : } Inspired from the earlier works \cite{5}, \cite{10}, we propose to concatenate the deep feature vectors of major knuckle, minor knuckle and nail to quantitatively evaluate the performance of cancelable finger dorsal biometrics. This test has been performed on PolyU FKI database with the same testing strategy. For perfect feature fusion, the real valued deep feature vectors of major knuckle, minor knuckle and nail are first normalized to some common range using min-max normalization technique and then fused using simplest sum rule. Subsequently, the deep fused feature vector is transformed to binary vector using the same criterion as discussed above in Section \ref{sec:trans}. Table \ref{tab:1x} illustrates the performance analysis of cancelable fused finger dorsal biometrics. The fusion performance shows superior recognition results with all bit map templates which outperforms the category wise performance of major knuckle, minor knuckle and nail. It implies that the fusion of finger components improves privacy protection compared to the corresponding systems based on a single modality. In other words, the binary hashed code generated by fused feature vector is more discriminative, not only in terms of recognition performance but it is more adversarial against hashing attacks. 
  
  \textbf{Security Analysis : } Let $\textbf{G}_\theta$ be the function approximated by FDFNet, that takes $I^i$ image corresponding to the subject $i$ and parameterized over $\theta$ to generate the feature $f^i$, as written below:
  \begin{equation}
  \textbf{G}_\theta(I^i)\rightarrow f^i
  \end{equation}
  Now, let us assume that the adversarial attacker somehow get hold of our network $\textbf{G}_\theta$. 
  Since network $\textbf{G}$ is known the reverse mapping network $\textbf{G}^\top$ architecture can be derived by symmetric transposition of $\textbf{G}_\theta$ network. However, the optimized network parameters $\delta$ of the reverse mapped network $\textbf{G}^\top_\delta$, (where $\delta \neq \theta$) could not be estimated or learned via. optimization techniques without having access to the input image and its corresponding feature vectors pair (even if actual $\theta$ is known). In literature \cite{rreverse}, it is well established that the complete recoverability of original input from the convolutional neural network got restricted due to the presence of non-linear activation functions like $Relu$ and $Maxpooling$ operations used in our architecture. However, complete recoverability of the original biometric image is necessary for any kind of adversarial attack because even ``very similar'' samples may led to completely different hashes. Thus no information, about the original biometric trait can be revealed. In such a scenario only brute force attack can work which is computationally expensive and infeasible. The proposed technique also satisfies the revocability condition by simply changing the user-specific key used in Bio-Hashing technique.

  \section{Conclusion and Future Scope} In this paper, we proposed a cancelable finger dorsal template generation technique based on deep-feature extraction using our novel architecture FDFNet and state-of-the-art Bio-Hashing technique. Our proposed approach shows quite good results on state-of-the-art knuckle databases namely PolyU FKP and PolyU Contactless FKI. Like the previous existing cancelable techniques our proposed architecture also offers same level of security to all the users enrolled in the database. But often in the organization structure different users can have different importance depending upon the cruciality of the project one is handling. So, for our future work, we propose to design a framework that incorporates user importance  along with its indigenous biometric-traits also. Possible future direction also includes testing our framework on other biometric traits like iris, fingerprints and face.

{\small
\bibliographystyle{ieee}
\bibliography{egbib}
}

\end{document}